\documentclass[10pt,twocolumn,letterpaper]{article}

\usepackage{iccv}
\usepackage{times}
\usepackage{epsfig}
\usepackage{graphicx}
\usepackage{amsmath}
\usepackage{amssymb}
\usepackage{booktabs}
\usepackage{multirow}
\usepackage{color}
\usepackage{bbm}

\usepackage[dvipsnames]{xcolor}
\usepackage[pagebackref,breaklinks,colorlinks,filecolor=blue, urlcolor=purple, citecolor=RoyalBlue]{hyperref}
\usepackage{subfloat}

 \iccvfinalcopy 

\def\iccvPaperID{5920} 

\ificcvfinal\fi

\begin{document}

\title{Toward Fine-Grained 3D Visual Grounding through Referring Textual Phrases}

\author{Zhihao Yuan$^{1,2, \dagger}$, Xu Yan$^{1,2, \dagger}$, Zhuo Li$^{1,2, \dagger}$,  Xuhao Li$^{1,2}$, Yao Guo$^{3}$, Shuguang Cui$^{2,1}$,  Zhen Li$^{2,1,}$\thanks{{ Corresponding author: Zhen Li. $^\dagger$ Equal first authorship.}} \\
$^{1}$FNii, CUHK-Shenzhen,  
	$^{2}$SSE, CUHK-Shenzhen,  
	$^{3}$Shanghai Jiao Tong University\\
	\url{https://yanx27.github.io/phraserefer/}
}

\maketitle
\ificcvfinal\thispagestyle{empty}\fi

\begin{abstract}
		Recent progress in 3D scene understanding has explored visual grounding (3DVG) to localize a target object through a language description.
		However, existing methods only consider the dependency between the entire sentence and the target object, ignoring fine-grained relationships between contexts and non-target ones. 
		In this paper, we extend 3DVG to a more fine-grained and interpretable task, called 3D Phrase Aware Grounding (\textbf{3DPAG}).
		The 3DPAG task aims to localize the target objects in a 3D scene by explicitly identifying all phrase-related objects and then conducting the reasoning according to contextual phrases.
		To tackle this problem, we manually labeled about {\textbf{227K phrase-level annotations}} using a self-developed platform, from 88K sentences of widely used 3DVG datasets, \ie, Nr3D, Sr3D and ScanRefer.
		By tapping on our datasets, we can extend previous 3DVG methods to the fine-grained phrase-aware scenario.
		It is achieved through the proposed novel phrase-object alignment optimization and phrase-specific pre-training, boosting conventional 3DVG performance as well.
		Extensive results confirm significant improvements, \ie, previous state-of-the-art method achieves {\textbf{3.9\%, 3.5\% and 4.6\%}} overall accuracy gains on Nr3D, Sr3D and ScanRefer respectively.
\end{abstract}

\begin{figure}[t]
		\begin{centering}
			\includegraphics[width=\linewidth]{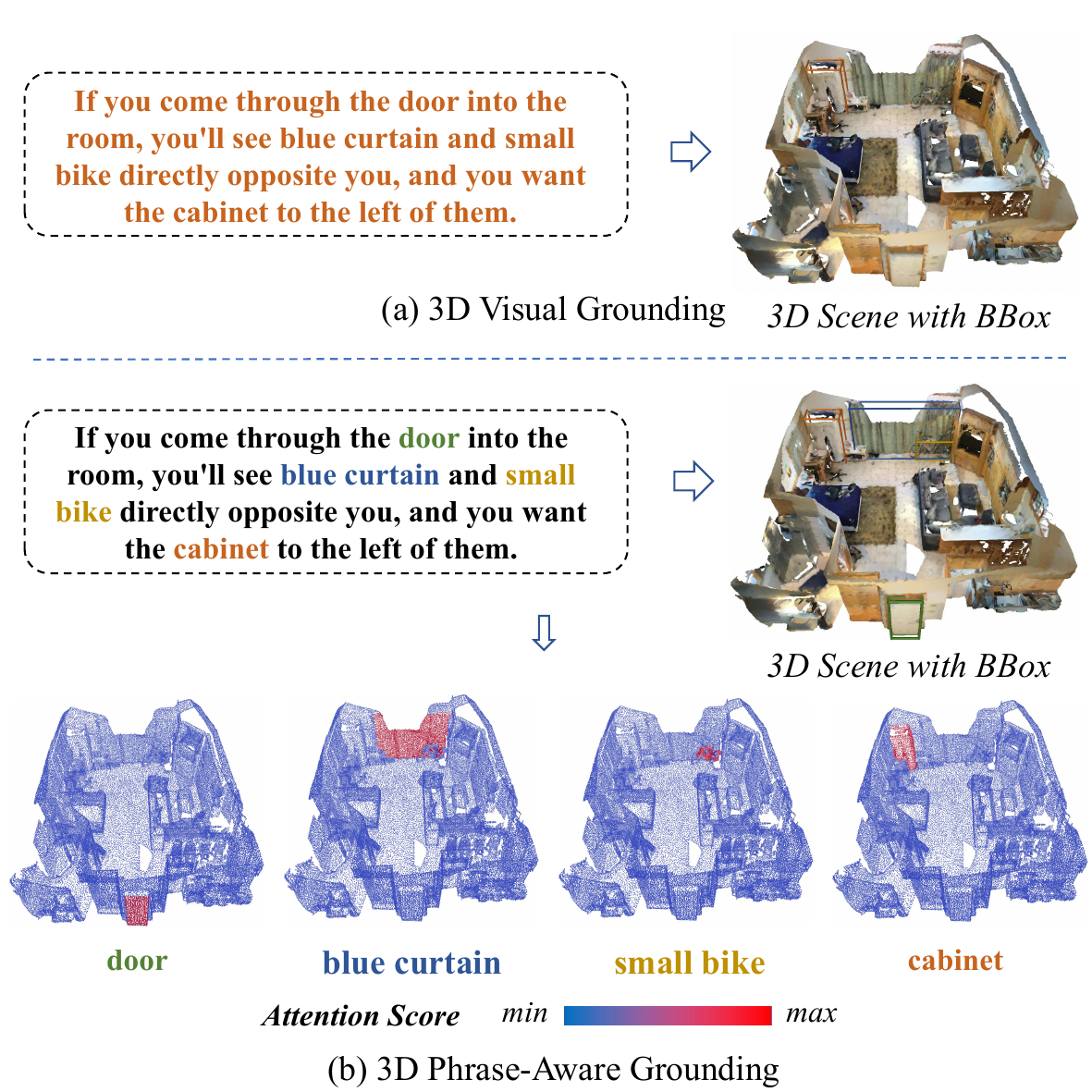}
			\caption{\textbf{3D Phrase-Aware Visual Grounding (3DPAG).} Compared with the previous 3DVG (a) only grounding the sole target object, 3DPAG (b) requires the neural listener to identify all phrase-aware objects (target and non-target) in the 3D scene and then explicitly conduct reasoning over all objects through the contexts. For the right part of (b), the 3D bounding boxes (BBox) are annotated using the same color as the corresponding object phrases in the sentence. For the bottom part of (b), we visualize the grounding attention score on the point cloud with respect to the given phrases based on our proposed method.}
			\label{fig:fig1}
		\end{centering}	
		\vspace{-0.3cm}
	\end{figure}
	
	\section{Introduction}
	\label{sec:intro}
	Hitherto, vision-language referring achieves significant progress in 2D computer vision community~\cite{kazemzadeh2014referitgame,mao2016generation,wang2018learning,liu2019improving,mogadala2019trends}, and has more and more applications for 3D scene understanding~\cite{chen2018text2shape,achlioptas2019shapeglot,chen2021scan2cap}.
	At the initial stage of 3D visual-language understanding, several approaches are proposed for {3D Visual Grounding} (3DVG) \cite{chen2020scanrefer,achlioptas2020referit3d,huang2021text,yuan2021instancerefer}, \ie, localizing a target object in the 3D scene based on an object-related description.
	However, 3DVG follows the mode of {\textbf{`one sentence to one object'}}, ignoring fine-grained cues of the 3D scene, \ie, multiple object entities are generally mentioned in the query sentence.
	Different from a limited number of objects in images for 2DVG, the object number in a real-world 3D scene is much larger and the spatial relationship is usually more complicated.
	As shown in Figure~\ref{fig:fig1}~(a), when a neural listener aims to localize the target ``cabinet'', it is difficult to simultaneously capture the ``door'' at a long-distance ``opposite'' position of the room.
	Existing methods learn the above relationship only through data-driven ways, \ie, implicitly exploiting positional encoding\cite{yang2021sat} or graph's edges\cite{yuan2021instancerefer}. Thus, they are difficult to capture fine-grained cues of non-target objects.
	Moreover, these methods cannot give an explainable result when the 3DVG model fails due to the inference error of the non-target objects.
	Hence, there is still a great demand for a more fine-grained 3DVG paradigm.

	In this work, we address the above problem following the principle of {\textbf{`one sentence to multiple objects'}}. 
	To this end, we propose the novel \textbf{Phrase-Aware 3D Grounding} (3DPAG) task as an extension of the existing 3DVG. 
	Compared with 3DVG only focusing on the described target object, 3DPAG also simultaneously localizes non-target objects in the 3D scene by referring to contextual phrases from the description.
	As illustrated in Figure~\ref{fig:fig1}~(b), 3DPAG requires the neural listener to localize all the objects mentioned in the sentence.
	Such a scheme enables the model to explore the complicated relationships between multiple objects and language much easier, thus achieving a more comprehensive understanding.
	To further facilitate the task, we extend the previous 3DVG datasets, \ie, ReferIt3D~\cite{achlioptas2020referit3d} and ScanRefer~\cite{chen2020scanrefer}, to their phrase-annotated versions.
	Specifically, we develop three new datasets and name them as {\textbf{\textit{Sr3D}++, \textit{Nr3D}++}} and {\textbf{\textit{ScanRefer}++}}, respectively.
	The former two are built upon ReferIt3D~\cite{achlioptas2020referit3d}, which contains 174.2K phrases annotations from 83.6K template sentences and 116.3K phrases annotations from 41.5K natural sentences, respectively.
	Similarly, \textit{ScanRefer}++ has 110.5K phrase annotations from 46.2K natural descriptions.
	
	Based on these newly annotated datasets, we modify previous approaches with proposed techniques and make them accessible to the more challenging 3DPAG task, benchmarking the new task as well.
	There are two plug-and-play techniques proposed: \textbf{1)}~Optimizing the {\textit{phrase-object alignment}} (POA) map, which is obtained from the multi-head cross-attention layer.
	\textbf{2)}~Conducting \textit{phrase-specific pre-training} strategy. Several phrase-specific masks alternate the target object to other mentioned objects, making the model easier to capture fine-grained contexts in the 3D scene.
	%
	%
	Experimental results confirm that adopting 3DPAG can effectively boost the performance of several existing 3DVG methods.
	%
	%
	By applying our proposed datasets and methodology, even state-of-the-art method~\cite{huang2022multi} gains great improvements on standard 3DVG task, \ie, achieving 59.0\%, 68.0\%, and 59.9\% accuracies on Nr3D, Sr3D, and ScanRefer datasets, respectively. 
	
	Our main contributions can be summarized as follows:
	\begin{itemize}
		\setlength{\itemsep}{0pt}
		\setlength{\parsep}{0pt}
		\setlength{\parskip}{0pt}
		\item[-] We introduce the phrase-aware 3D visual grounding (3DPAG) task based on the fine-grained 3DVG datasets, containing about 227K newly phrase-level manual annotations from 88K sentences.
		\item[-] We propose novel techniques, named phrase-object alignment map optimization and phrase-specific pre-training. They make previous 3DVG models accessible to the more challenging 3DPAG task.
		\item[-] The developed dataset and components confirm that the 3DPAG greatly improves the 3DVG performance, and the enhanced approach achieves state-of-the-art performance on several benchmarks.
		
	\end{itemize}

	\section{Related Work}
	\label{sec:related}
	\subsection{3D Visual Grounding}
	3D visual grounding (3DVG) is an emerging research field. 
	ScanRefer~\cite{chen2020scanrefer} and ReferIt3D~\cite{achlioptas2020referit3d} are pioneers for the 3DVG, in which both datasets and baseline methods are concurrently proposed.
	%
	%
	Previous studies~\cite{chen2020scanrefer,achlioptas2020referit3d,huang2021text,yuan2021instancerefer} formulate 3DVG as a matching problem. 
	They first generate multiple 3D object proposals through either ground truths~\cite{achlioptas2020referit3d} or a 3D object detector~\cite{chen2020scanrefer}, and then fuse 3D proposals with the language embeddings to predict proposals’ similarity scores. 
	%
	The proposal with the highest score will be selected as the target object.
	Recent works \cite{roh2021languagerefer,yang2021sat,zhao20213dvg,jain2022bottom} use Transformer~\cite{vaswani2017attention} architecture to encode the relationship between language and objects' proposals, where \cite{yang2021sat} and \cite{roh2021languagerefer} exploit additional 2D images and orientation annotation during the training stage to boost the performance, respectively.
	Recently, \cite{zhenyu2021d3net,cai_3djcg_2022} combine the task of 3DVG with the task of 3D dense captioning to mutually enhance the performance of the two tasks.
	MVT~\cite{huang2022multi} explores multi-view information in the 3D scene to make a more reliable grounding.
	Nevertheless, all of these methods exploit the entire sentence to identify a single target object. Such a scheme overlooks the fine-grained relationships between the language and all objects, especially non-target objects, making the grounding process unexplainable as well.
	
	\begin{figure*}[t]
		\begin{centering}
			\includegraphics[width=\textwidth]{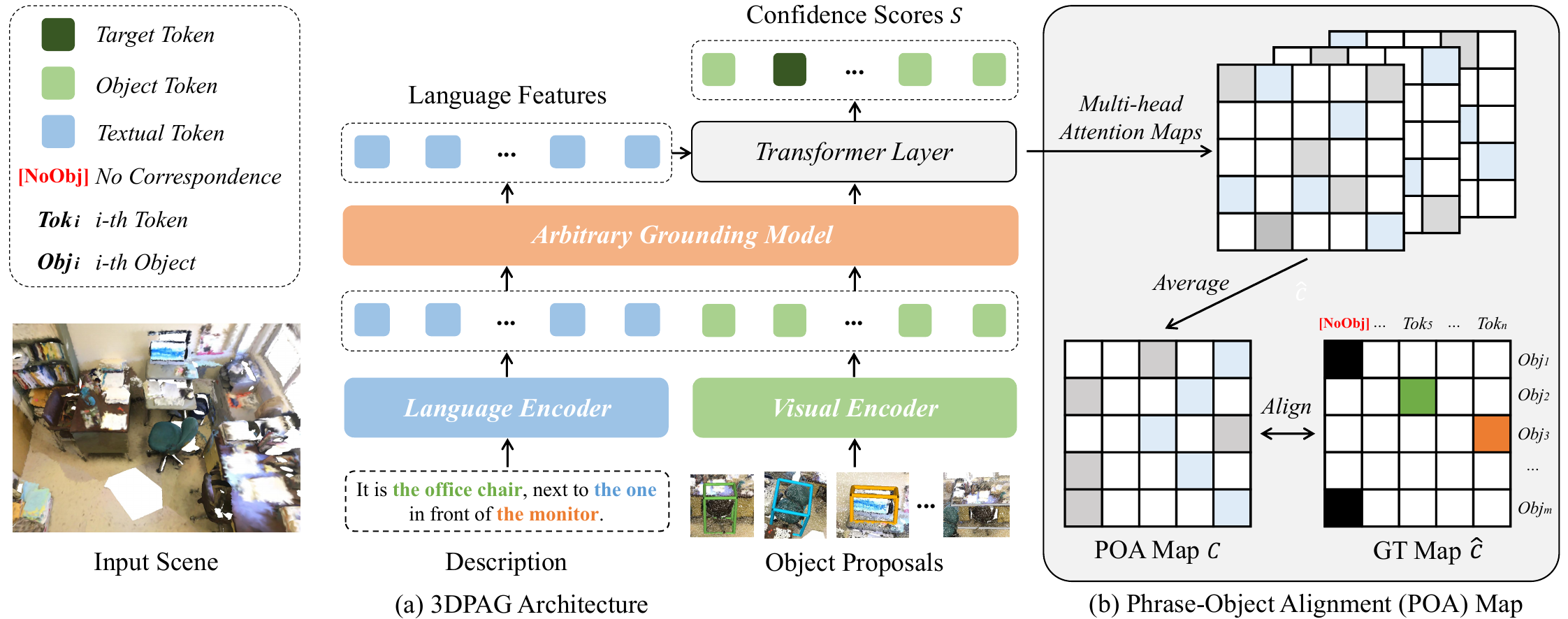}
			\vspace{-.3cm}
			\caption{\textbf{3D phrase aware grounding (3DPAG) architecture with Phrase-Object Alignment (POA) Optimization.} Part (a) illustrates the 3DPAG architecture, and part (b) shows the process of optimizing the phrase-object alignment map.}
			\label{fig:fig4}
		\end{centering}	
	\end{figure*}
	
	\subsection{Phrase Grounding on Images}
	Different from visual grounding which adopts the whole sentence to localize the target object on images~\cite{wang2018learning}, phrase grounding is defined as spatial localization with a textual phrase.
	Existing approaches exploit different techniques to realize phrase grounding on 2D images, where some of them map the visual and textual modalities onto the same feature space~\cite{plummer2017phrase,wang2018learning}, and the others aim to reconstruct the corresponding phrase~\cite{rohrbach2016grounding}. 
	For the development of architecture, Hinami \& Satoh~\cite{hinami2017discriminative} train the object detectors with an open vocabulary, Plummer~\etal~\cite{plummer2018conditional} condition the textual representation on the phrase category. 
	Recent approaches integrate object proposals with Transformer models~\cite{li2019visualbert,lu2019vilbert}.
	Besides, some previous works focus on utilizing unsupervised~\cite{rohrbach2016grounding}, weakly supervised~\cite{zhao2018weakly,xiao2017weakly,chen2018knowledge} manners to achieve phrase grounding.
	
	Different from image-based phrase grounding, our 3D phrase aware grounding (3DPAG) has the following properties: 
	\textbf{1)} In the dataset aspect, sentences used in 2D phrase grounding are generally captions for the entire scene. In contrast, since our datasets are initiated for visual grounding, the majority of sentences are descriptions of the target object.
	\textbf{2)} In the methodology aspect, approaches in 2D phrase grounding only take a textual phrase and the corresponding image as inputs. By contrast, we utilize the entire sentence and the 3D scene as inputs to localize the target object, while identifying all the phrase-related objects mentioned in the context is an auxiliary task.

	\subsection{Vision-Language Pre-training}
	Vision-language pre-training has become a hotspot in cross-modal learning \cite{tan2019lxmert,li2020unimo,jia2021scaling,ramesh2021zero}. 
	CLIP~\cite{radford2021learning} uses independent encoders to extract features of language and vision separately and then aligns textual and visual information into a joint semantic space with a contrastive criterion. 
	Existing methods usually pre-train transformer models on large-scale image-text paired datasets~\cite{lin2014microsoft,krishna2017visual} and fine-tune on downstream tasks. 
	These approaches have achieved state-of-the-art results across various multi-modal tasks such as visual question answering~\cite{li2021semvlp}, image captioning~\cite{li2020oscar}, phrase grounding~\cite{kamath2021mdetr} and image-text retrieval~\cite{zhang2021vinvl}. 
	Nevertheless, due to the limited number of available 3D data and the difficulty of 3D representation learning, there is no pre-training study for 3DVG till now.
	%

	
	\begin{figure*}[t]
		\begin{centering}
			\includegraphics[width=\textwidth]{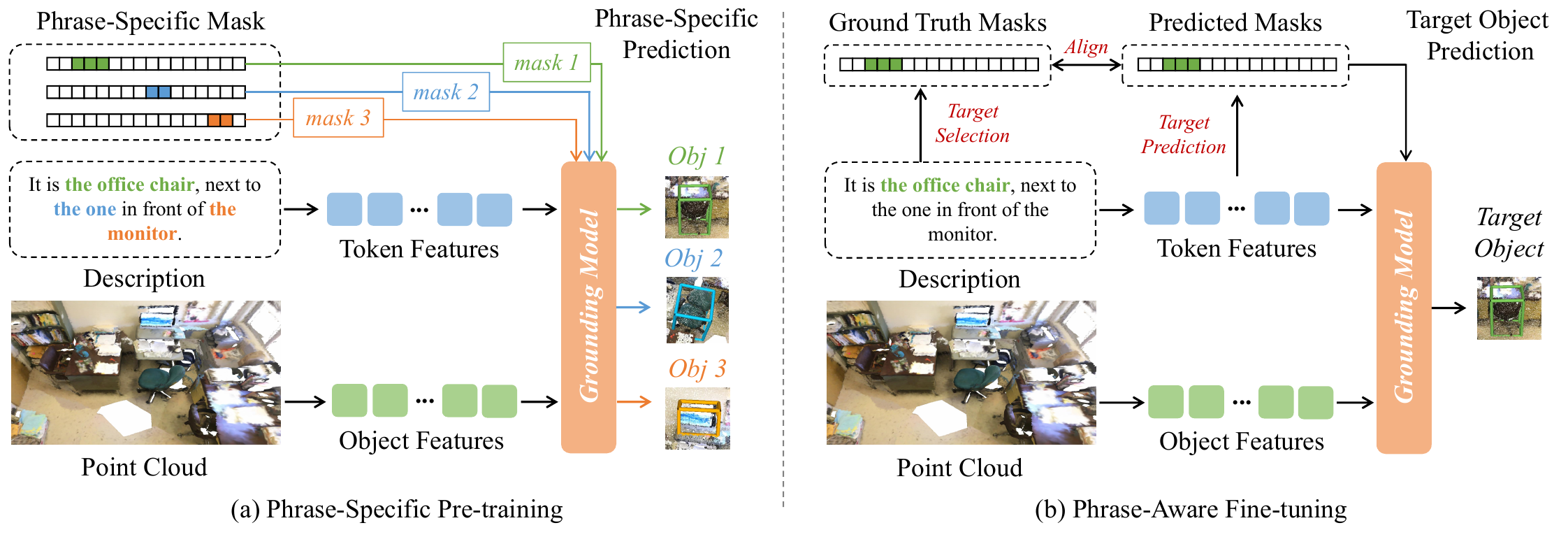}
			\caption{\textbf{Phrase-Specific Training.} In the phrase-specific pre-training stage, we generate the phrase-specific masks according to the grounding truth phrases, \eg, setting the position of ``the office chair'' to $1$ and other positions to $0$. Then, we design the network to predict the corresponding object of the selected phrase. During the fine-tuning stage, we only predict the target object referred in the sentence.}
			\label{fig:fig3}
		\end{centering}	
	\end{figure*} 
	
	\section{Problem Statement}
	\label{sec:problem}
	The problem of 3DVG is to localize a specific object in a 3D scene using natural language.
	%
	In line with previous studies, the 3D scene is represented by object proposals $O$, which can be extracted either through ground truth objects~\cite{achlioptas2020referit3d} or a 3D object detector~\cite{chen2020scanrefer}.
	Language query with token length $L$ is embedded to word features $E = \{e\}_{i=1}^L \in \mathbb{R}^{L\times D_e}$ by the pre-trained text encoder, \eg, BERT \cite{devlin2018bert}.
	Previous 3DVG approaches conduct cross-modal fusion between $O$ and $E$ to generate confidence scores for each object:
	\begin{align}
	\text{Enc}_S(\text{Scene})& \mapsto O,~~\text{Enc}_Q(\text{Query}) \mapsto E,\\ &\text{3DVG}(O,E) \mapsto S,
	\label{3DVG}
	\end{align}
	where $O = \{o\}_{i=1}^M \in \mathbb{R}^{M\times D_{obj}}$ are $M$ object features with $D_{obj}$ dimensions, and $S \in \mathbb{R}^M$ are confidence scores of $M$ objects. The proposal that has the highest score will be selected as the target object.
	%
	
	%
	
	Since previous works only focus on localizing a single target object, other non-target objects mentioned in the input sentence are overlooked.
	Such information about all objects is important for humans to identify the target objects, but it is hard for neural networks to understand without direct supervision.
	To address this problem, we propose a task, namely 3D phrase-aware grounding (3DPAG), that grounding the non-target objects along with the target one:
	\begin{align}
	\text{3DPAG}(O,E) &\mapsto  (S, C).
	\label{3DPAG}
	\end{align}
	To achieve this goal, we simultaneously predict a phrase-object alignment (POA) map $C \in \mathbb{R}^{M\times (L+1)}$ together with confidence scores $S$.
	In a POA map, the element $C_{[i, j]}$ in the $i$-th row and $j$-th column represents the alignment score between $i$-th object proposal and $j$-th language token, as shown in Figure~\ref{fig:fig4}(b).
	Since not all object proposals are mentioned in the sentence, we set an extra [NoObj] token besides the original $L$ tokens to represent the proposal without the corresponding phrase.
	
	In this manner, we can not only obtain the target object by ranking the confidence scores $S$, but also gain fine-grained correspondence of non-target ones through the proposed POA map.
	In practice, if we want to find out an object related to a certain token, we can rank the corresponding column and select the object that has the highest score.
	If there are multiple tokens describing the same object, we average the scores of these columns before ranking them.

	\section{Methodology}
	\label{sec:method}
	In this section, we introduce our proposed components for 3DPAG.
	We introduce our baseline in Sec.~\ref{sec:baseline}.
	In Sec.~\ref{sec:poc}, we demonstrate the phrase-object alignment (POA) map optimization, which makes the grounding more fine-grained and effective.
	In Sec.~\ref{sec:pre-training}, we design a phrase-specific pre-training, which utilizes the phrase annotations and fine-grained cues, further boosting the performance.

	\subsection{3D Visual Grounding Baseline}
	\label{sec:baseline}
	Figure~\ref{fig:fig4}~(a) illustrates how we can make a typical architecture for 3D visual grounding accessible to 3D phrase aware grounding (3DPAG).
	Generally, the network consists of two encoders to independently deal with the object proposals and language descriptions.
	On the one hand, for the visual encoder, the global features for objects are extracted by a feature extractor such as PointNet++~\cite{qi2017pointnet++}. Then the fine-grained classes of the global features are predicted.
	Afterward, the position and size information, \ie, center and bounding box, are embedded into a positional encoding vector through linear transformation, which is subsequently concatenated with global features.
	On the other hand, the language encoder directly embeds the text input with a pre-trained BERT model~\cite{devlin2018bert}, where an extra [CLS] token is set to predict the target category through the input sentence. 
	After that, vision and language tokens are fed into a 3DVG model, obtaining updated object and language features, respectively.
	In practice, previous approaches design different 3DVG models through diverse architectures, such as Transformer~\cite{yang2021sat} and language-guided GCN~\cite{yuan2021instancerefer}. 
	All of these methods can be directly adopted in our architecture.
	Moreover, we add an extra cross-attention layer~\cite{vaswani2017attention} to fuse the language and object predictions. Thus the obtained attention maps can be used to predict the subsequently introduced phrase-object alignment (POA) map.
	Finally, the output features are fed into two fully connected layers, producing a set of confidence scores.
	The object proposal with the highest grounding score is selected as the final grounding prediction.

	\begin{table*}[t]
		\centering
		\caption{The statistics of the developed datasets for 3D phrase aware grounding, \ie, Nr3D++, Sr3D++ and ScanRefer++.}
		\begin{tabular}{l|cc|cc|cc}
			\toprule[1pt]
			\multicolumn{1}{l|}{\multirow{2}[0]{*}{Dataset}} & \multicolumn{2}{c|}{Nr3D++} & \multicolumn{2}{c|}{Sr3D++} & \multicolumn{2}{c}{ScanRefer++} \\
			& \multicolumn{1}{c}{train} & \multicolumn{1}{c|}{val} & \multicolumn{1}{c}{train} & \multicolumn{1}{c|}{val} & \multicolumn{1}{c}{train} & \multicolumn{1}{c}{val} \\\hline
			Number of sentences & ~32,919 & 8,584~ & 65,846 & 17,726~ & ~36,665 & 9,508~  \\
			Number of phrases & ~92,691 & 23,620~ & ~137,158 & 37,124~& ~87,391 & 23,103~  \\
			Number of phrases per sentence~~~ & 2.82     & 2.75     & 2.08     & 2.09     & 2.38     & 2.43 \\
			Average length of phrases & 2.34 &2.37 &1.17 &1.17 &1.76  &1.77  \\
			\bottomrule[1pt]
		\end{tabular}%
		\label{tab:statistic}%
	\end{table*}

	\subsection{Phrase-Object Alignment Optimization}
	\label{sec:poc}
	During the training phase, the phrase-object alignment (POA) map is not predicted directly as depicted in Eqn.~\eqref{3DPAG}.
	Intuitively, we find that the attention map in the last cross-attention layer is a nature POA map, which can be directly optimized.
	As shown in Figure~\ref{fig:fig4}~(b), we extract the multi-head attention maps from the last cross-attention layer and average them along the head dimension.
	It shares the same property as the POA map described in Eqn.~\eqref{3DPAG}.
	The ground truth (GT) map in Figure~\ref{fig:fig4}~(b) can be generated by phrase annotations.
	Assume there are $M$ object proposals and $L$ language tokens, the GT map $\hat{C}$ should have the shape of $M\times (L+1)$, where the additional dimension in the column represents [NoObj] correspondence.
	
	For each object proposal (\ie, the row of the map), if it is mentioned by any token, the column of the corresponding token should be 1 (0 for other columns).
	If there is no correspondence, the column of [NoObj] is set to 1.
	For example, for the last two tokens, ``the monitor'' in Figure~\ref{fig:fig4}, which correspond to the third object, the last two columns in the third row should be set to 1.
	After that, we optimize the POA map using a multi-class cross entropy (CE) loss.
	During the inference, the POA map conducts a fine-grained 3DVG. If one wants to know which object corresponds to ``the monitor'' in Figure~\ref{fig:fig4}, the score for each object can be gained by averaging the last two columns and the object that has the highest score is the desired output. 
	
	%

	\subsection{Phrase-Specific Training}
	\label{sec:pre-training}
	Except for utilizing phrase information in the training stage, we also propose a phrase-specific pre-training strategy, which is demonstrated in Figure~\ref{fig:fig3}.
	The pre-training strategy is trained only on the 3DVG task but can be applied to both 3DVG and 3DPAG models.
	Specifically, we manually set several phrase-specific masks according to the number of objects mentioned in the sentence for pre-training.
	Suppose there are $K$ phrases, represented as $P = \{p_i\}_{i=1}^K$, in a query sentence, we define their phrase-specific masks $G = \{g_i\}_{i=1}^K \in \mathbbm{1}^{K\times L}$ as follows:
	\begin{align}
	g_{i,j} = 1\ {\text{if}}~\ j\text{-th token} \in p_i\ {\text{else}}\ 0,
	\end{align}
	where each $g_i \in G$ is a phrase-specific mask corresponding to the phrase $p_i$.
	Taking Figure~\ref{fig:fig3}~(a) as an example, since there are three objects mentioned in the sentence, \ie, ``the office chair'', ``the one'' (another chair) and ``the monitor'', we set three masks ($K=3$) whose length are equal to the number of tokens in the sentence.
	For each mask, only positions that are related to the specific object are equal to 1.
	After that, we use these masks as extra positional encoding and concatenate them with language features $E$.
	In each iteration, we randomly choose one mask $g_i \in G$ and object features $O$ to identify the object referred by $p_i$ via
	\begin{align}
	E \oplus g_i \mapsto E',\ \text{3DVG}(O&,E') \mapsto S(p_i),
	\end{align}
	where $S(p_i)$ means the model generates confidence scores for the object related to the phrase $p_i$.
	Through such a manner, fine-grained information can be exploited, making the model easier to capture non-target objects.
	
	After pre-training, we fine-tune our model on either 3DVG or 3DPAG tasks, as shown in Figure~\ref{fig:fig3}~(b). 
	Specifically, we add a linear layer to predict the mask that is only related to the target object, supervised by the ground truth.
	%

	\section{Dataset}
	To extend 3D visual grounding (3DVG) to 3D phrase-aware grounding (3DPAG), another contribution of this paper lies in the development of 3DVG datasets with extra phrase-level annotations.
	In this section, we first introduce the original 3DVG datasets in Sec.~\ref{3dvgdata}. 
	Then, the data labeling process and statistics of the proposed 3DPAG datasets are demonstrated in Sec.~\ref{3pagdata}.

	\subsection{3D Visual Grounding Datasets}
	\label{3dvgdata}
	There are two existing datasets for 3DVG, \ie, ScanRefer~\cite{chen2020scanrefer} and ReferIt3D~\cite{achlioptas2020referit3d}.
	
	\begin{itemize}
		\setlength{\itemsep}{0pt}
		\setlength{\parsep}{0pt}
		\setlength{\parskip}{0pt}
		\item[-] \textbf{ScanRefer}~\cite{chen2020scanrefer} consists of 51.5K sentence descriptions for 800 ScanNet scenes \cite{scannet}. 
		The dataset is split into 36,655 samples for training, 9,508 samples for validation, and 5,410 samples for testing, respectively.
		
		\item[-] \textbf{ReferIt3D}~\cite{achlioptas2020referit3d} uses the same train/valid split with {ScanRefer} on ScanNet and contains two sub-datasets, where \textbf{Sr3D} (Spatial Reference in 3D) has 83.5K synthetic expressions generated by templates and \textbf{Nr3D} (Natural Reference in 3D) consists of 41.5K human annotations collected similarly as ReferItGame~\cite{kazemzadeh2014referitgame}. 
		
	\end{itemize}

	\subsection{3D Phrase Aware Grounding Datasets}
	\label{3pagdata}

	\noindent\textbf{Data Annotations.}
	Since the Sr3D dataset is generated from language templates, we can acquire the corresponding phrase annotations.
	For the other two datasets, we deploy a web-based annotation interface modified by ScanRefer Browser\footnote{\url{github.com/daveredrum/ScanRefer_Browser}} on Amazon Mechanical Turk (AMT) to identify objects' phrases. 
	During the annotation process, annotators need to identify the phrases in the original sentence and find out the corresponding object in the 3D scene.
	
	The 3D web-based UI allows users to select arbitrary texts, and it will automatically save the phrase by recording the start and end positions.
	When the annotators search for an object, the selected object is highlighted while others are faded out. Meanwhile, a set of captured image frames are shown to compensate for incomplete details of the 3D reconstruction.
	The position and viewpoint of the virtual camera are allowed for adjusting flexibility to examine the target object better.
	To ensure the annotation quality, we allow users to select a ``not make sure'' button during the annotation, and these suspicious sentences will go through a double-check by us.
	Each sentence will be labeled by two annotators. 
	After labeling, the annotators will verify the results according to each other's annotations.
	When labeling ScanRefer dataset, it takes an average of 40 minutes for an annotator to complete a scene. There are \textbf{3664} man-hours to complete the whole labeling.

	\begin{table}[t]
		\caption{\textbf{3DVG-GT on Nr3D, Sr3D and ScanRefer datasets}. We compare them by overall accuracy.}
		
		\setlength\tabcolsep{0.5pt}
		\centering 
		\small
		\resizebox{\linewidth}{!}{
			\begin{tabular}{cccc}
				\toprule[1pt]
				Method & Nr3D & Sr3D & ScanRefer \\
				\hline
				ScanRefer \cite{chen2020scanrefer} & $34.2 $ & - & 44.5 \\
				ReferIt3D \cite{achlioptas2020referit3d} & $35.6 $ & 40.8 & 46.9 \\
				TGNN~\cite{huang2021text} & 37.6 & 45.2 & - \\
				InstanceRefer \cite{yuan2021instancerefer} & $38.8 $ & 48.0 & 49.2 \\
				3DVG-Trans. \cite{zhao20213dvg} & 40.8 & 51.4 & -\\
				FFL-3DOG \cite{feng2021free} & $41.7 $ & - & -  \\
				TransRefer3D~\cite{he2021transrefer3d} & 42.1 & 57.4 & - \\
				LanguageRefer \cite{roh2021languagerefer} & 43.9 & 56.0 & - \\
				BUTD-DETR \cite{jain2022bottom} & 54.6 & 67.0 & - \\
				SAT~\cite{yang2021sat} & 49.2 & 57.9 & 53.8 \\
				MVT~\cite{huang2022multi} & 55.1 & 64.5 & 55.3 \\
				\hline
				\textbf{SAT + Ours} & ~~~~~~~$\mathbf{54.4}_{\color{RoyalBlue}{+5.2}}$    & ~~~~~~~$\mathbf{63.0}_{\color{RoyalBlue}{+5.1}}$ & ~~~~~~~$\mathbf{57.5}_{\color{RoyalBlue}{+3.7}}$ \\
				\textbf{MVT + Ours} & ~~~~~~~$\mathbf{59.0}_{\color{RoyalBlue}{+3.9}}$   & ~~~~~~~$\mathbf{68.0}_{\color{RoyalBlue}{+3.5}}$  & ~~~~~~~$\mathbf{59.9}_{\color{RoyalBlue}{+4.6}} $ \\
				
				\bottomrule[1pt]
			\end{tabular}
		}
		\label{tab:nr3d}
	\end{table}

	\noindent\textbf{Data Statistics.}
	Thanks to the high-quality data annotations, we can extend the previous 3DVG datasets to phrase-aware versions, \ie, Nr3D++, Sr3D++, and ScanRefer++.
	As shown in Table~\ref{tab:statistic}, among three 3DPAG datasets, Sr3D++ has the largest number of sentences and phrases since it is originally generated by machine templates.
	Nr3D++ has the smallest number of sentences, but the number of phrases per sentence achieves 2.82 in the training split, surpassing that of Sr3D++ (2.08) and ScanRefer++ (2.38).
	Also, the average length of phrases in Nr3D++ is the longest.

	\section{Experiments}
	\label{sec:exp}
	
	\subsection{Experiment settings}

	\noindent\textbf{Baselines.}  
	There are several evaluation modes in 3DVG. The 3DVG models are tested with (1) detector-generated proposals (\ie, \textbf{3DVG-Det}); and (2) the ground truth proposals (\ie, \textbf{3DVG-GT}) to prevent the experimental bias via a more powerful detector.
	We rigorously benchmark the baseline and enhanced model in this paper and tested them in both modes.
	Foremost, we follow the settings~\cite{achlioptas2020referit3d,yang2021sat,roh2021languagerefer} of using the ground truth proposals on each dataset, where SAT~\cite{yang2021sat} and MVT~\cite{huang2022multi} are used as our baseline.
	After that, we evaluate models with detector-generated proposals on ScanRefer dataset, spanning different detectors, \eg, SAT~\cite{yang2021sat} with VoteNet~\cite{qi2019deep} and D3Net~\cite{zhenyu2021d3net} with more powerful PointGroup~\cite{jiang2020pointgroup}.
	The comprehensive experiments show our proposed datasets and methods significantly improve 3DVG.

	\begin{table}[t]
		\caption{\textbf{3DVG-Det on ScanRefer dataset.} We report the result by IoU@0.5 metric. `VN', `PG', `GF' represent VoteNet~\cite{qi2019deep}, PointGroup~\cite{jiang2020pointgroup}, and GroupFree3D \cite{liu2021group} respectively. (*) Results without semi-supervised training on \textbf{extra data}.}
		\footnotesize
		\centering
		\resizebox{\linewidth}{!}{
			\begin{tabular}{ccccc}
				\toprule[1pt]
				Method & Detector & Uni.  & Mul. & Overall \\
				\hline
				ScanRefer~\cite{chen2020scanrefer}& VN & 53.51 & 21.11 & 27.40 \\
				TGNN~\cite{huang2021text}& PG & 56.80 & 23.18 & 29.70 \\
				InsRefer~\cite{yuan2021instancerefer}& PG & 66.83 & 24.77 & 32.93 \\
				FFL-3DOG~\cite{feng2021free} & VN & 67.94 & 25.70 & 34.01 \\
				3DVG-Trans.~\cite{zhao20213dvg} & Other & 60.64 & 28.42 & 34.67 \\
				SAT~\cite{yang2021sat} & VN   & 50.83  & 25.16  & 30.14 \\
				D3Net*~\cite{zhenyu2021d3net} & PG& 71.04 & 27.40 & 35.62 \\
                    B-DETR \cite{jain2022bottom} & GF & 66.30 & 35.10 & 39.80 \\
				\hline
				~\textbf{SAT + Ours} & VN & 58.21 & 26.76 & ${33.01}_{\color{RoyalBlue}{+2.87}} $ \\
				~\textbf{D3Net + Ours} & PG & \textbf{72.50}  & 28.84 & $37.30_{\color{RoyalBlue}{+1.68}} $\\
                    ~\textbf{B-DETR + Ours} & GF & 67.16 & \textbf{37.05} &  $\mathbf{41.50}_{\color{RoyalBlue}{+1.70}}$ \\
				\bottomrule[1pt]
			\end{tabular}
		}
		\label{tab:scanrefer}
	\end{table}
	
	\begin{figure*}[t]
		\begin{centering}
			\includegraphics[width=0.9\textwidth]{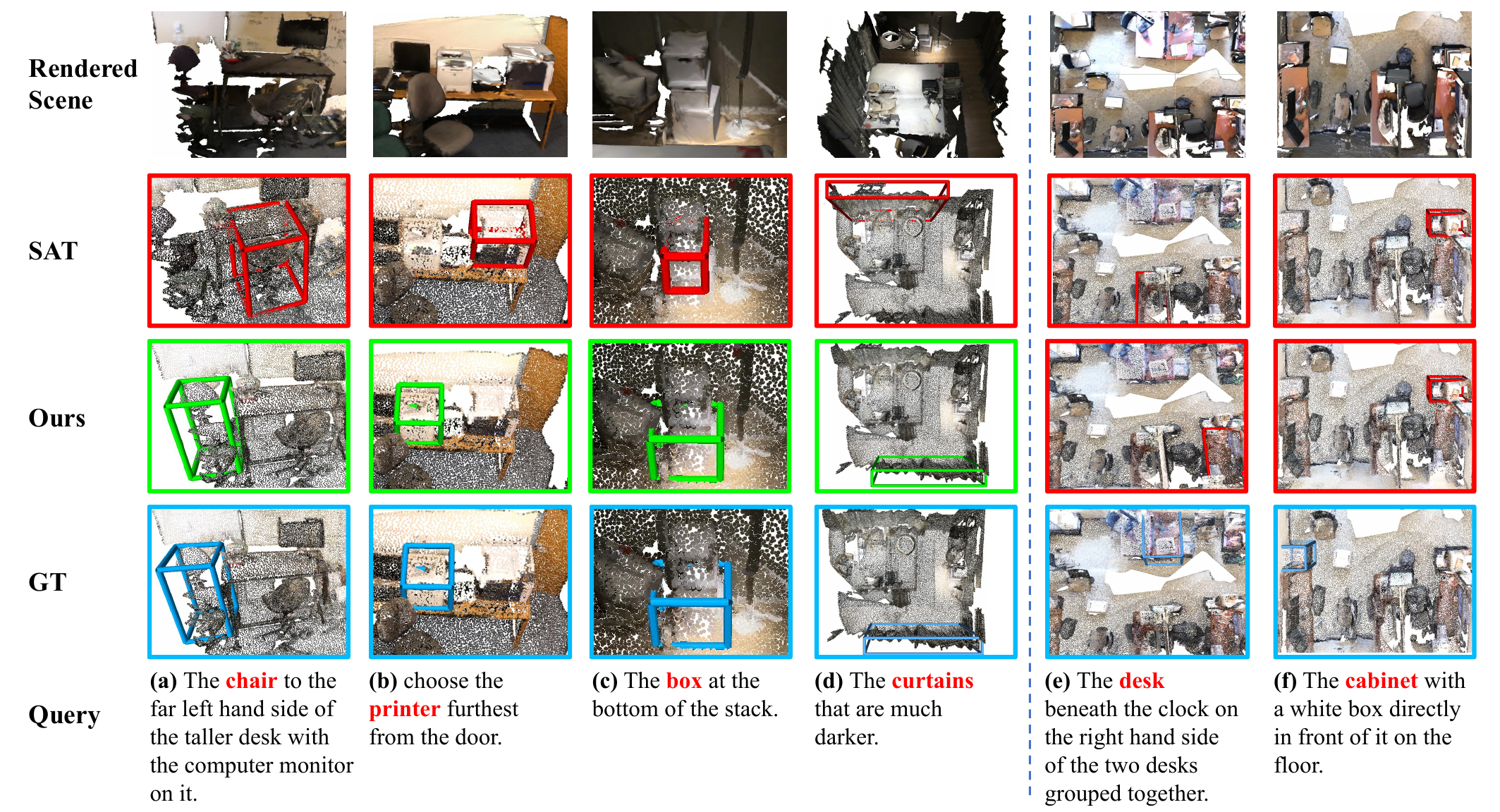}
			\caption{\textbf{Visualization Results of 3DVG.} We visualize the visual grounding results of SAT and ours. The four left examples are our correct predictions, while SAT failed. The two right examples show the representative failures for both the baseline and our method. The green/red/blue colors illustrate the correct/incorrect/GT boxes. The target class for each query is shown in red color. We provide rendered scenes in the first row for better visualization. Best viewed in color.}
			\label{fig:fig5}
		\end{centering}	
	\end{figure*}
	
	\noindent\textbf{Evaluation Metric.}  In existing 3DVG-GT, models are evaluated by the overall accuracy $Acc_{{\text{VG}}}$, \ie, whether the model correctly selects the target object among $M$ proposals given a language query. 
	For the mode of 3DVG-Det, it calculates the 3D intersection over union (IoU) between the predicted bounding box and ground truth. The Acc@0.5IoU is adopted as the evaluation metric.
	Moreover, the accuracy is reported in ``unique''(19\%) and ``multiple'' (81\%) categories, respectively. 
	
	In this paper, we also design a novel metric $Acc_{{\text{PAG}}}$ for the 3DPAG task.
	Specifically, for each sentence, we simultaneously evaluate the accuracy of a target object and mentioned non-target objects, and the final result is gained by multiplication of two accuracies: 
	\begin{align}
	\label{Acc_PAG}
	Acc_{{\text{PAG}}} = Acc_{\text{VG}} \times Acc_{{\text{non-target}}}. 
	\end{align}
	For each sentence, $Acc_{{\text{VG}}}$ is 1 if the target object is identified correctly, otherwise 0.
	$Acc_{{\text{non-target}}}$ is the grounding accuracy of non-target but mentioned objects (excluding the target one).
        If multiple objects are related by a phrase, we allow the model to choose an arbitrary one (less than 0.1\%).
	The $Acc_{{\text{PAG}}}$ on the whole dataset is the averaged accuracy of all sentences, and thus it can be treated as the weighted version of $Acc_{{\text{VG}}}$.
	By using this weighted accuracy, we can prevent the model from only predicting the right result through learning data distribution, \ie, $Acc_{{\text{PAG}}}$ equals to 1 only if the model correctly identifies the target object and all non-target objects.

	\noindent\textbf{Implementation Details.}
	%
	%
	%
	%
	During the pre-training stage, the model is trained with the Adam optimizer with a batch size of 16. We set an initial learning rate of $10^{-4}$ and reduce the learning rate by a multiplicative factor of $0.65$ every 10 epochs for a total of 100 epochs.
	During the fine-tuning stage, we reduce the initial learning rate to $10^{-5}$ and keep other hyperparameters the same.
	%

	\subsection{3D Visual Grounding Results}
	\noindent\textbf{3DVG-GT.}
	The results of 3D visual grounding (3DVG) with ground truth proposals are shown in Table~\ref{tab:nr3d}.
	By effectively utilizing fine-grained phrase annotations in both pre-training and training processes, our proposed dataset and components improve the SoTA method MVT~\cite{huang2022multi}, from 55.1/64.5/55.3\% to 59.0/68.0/59.9\% on Nr3D/Sr3D/ScanRefer datasets in terms of overall accuracy, respectively. 
	Moreover, there is a considerable improvement on the other baseline SAT~\cite{yang2021sat} with lower initial performance. Our dataset and proposed components boost its performance by 5.2\%, 5.1\%, and 3.7\%, respectively. 
	The performance boosts on 3DVG-GT are more reliable, since it only considers the referring ability of the model and ignores the effect of different detection architectures.
	
	\noindent\textbf{3DVG-Det.}
	3D visual grounding with detected proposals is a more complicated task since it also considers detection results.
	Table~\ref{tab:scanrefer} reports the 3DVG on the ScanRefer validation set~\cite{chen2020scanrefer} with detected object proposals. 
	After equipping our POA optimization and pre-training strategy, the enhanced model achieves an accuracy of 41.5\%, outperforming the previous state-of-the-arts~\cite{zhenyu2021d3net,yang2021sat,jain2022bottom}. 
	Moreover, we find out that our proposed components and dataset work well across different object detectors, \ie, VoteNet~\cite{qi2019deep}, PointGroup~\cite{jiang2020pointgroup} and GroupFree~\cite{liu2021group}, which further shows the effectiveness.
	
	\noindent\textbf{Visualization.} In Figure~\ref{fig:fig5}, we show some grounding predictions on Nr3D with SAT baseline. 
	As shown in examples (a-d), our method corrects failure cases made by baseline. 
	For instance (a, b), our method correctly understands the relationships between mentioned objects ``chair'' and ``monitor'', ``printer'' and ``door''. 
	Besides, our method also distinguishes the phrases related to the spatial position, \eg, ``far left'' and ``furthest''. 
	In (c, d), our method reasons the target from multiple similar objects by their position and color information, \ie, ``bottom'' and ``darker''. 
	The (e, f) are failure cases, which are caused by severe distractors of the same classes in a complicated scene.

	\subsection{3D Phrase Aware Grounding Results}
	\noindent\textbf{Fully supervised 3DPAG-GT.}
	We extend previous 3DVG methods to the newly proposed 3DPAG task by optimizing POA maps, as depicted in Sec.~\ref{sec:baseline}, whose attention maps are used for phrase-object alignment map generation.
	The fully supervised 3DPAG results are shown in the upper part of Table~\ref{tab:3dpag}. Besides computing $Acc_{\text{PAG}}$ introduced in Eqn. \ref{Acc_PAG} (\textbf{PAG}), we also provide the overall accuracy $Acc_{\text{PG}}$ of all phrases (\textbf{PG}).
	
	As shown in Table~\ref{tab:3dpag}, MVT equipping our pre-training strategy (\ie, PSP) achieves the highest results in all metrics, especially on ScanRefer++ dataset which contains more ambiguous phrases as distractors. 
	Specifically, it respectively improves $Acc_{\text{PAG}}$ and $Acc_{\text{PG}}$ by \textbf{7.4\%} and \textbf{6.0\%} compared with naive extension of MVT. 
	Moreover, after phrase-specific pre-training (PSP) with phrase-level annotations, SAT also significantly improves the performance for 3DPAG, increasing 8.6\% and 6.2\% for PAG and PG.

	%
	
	\begin{table}[t]
		\centering
		\caption{\textbf{Benchmarking 3DPAG.} PAG means phrase aware grounding accuracy $Acc_{\text{PAG}}$ and PG means overall phrase grounding accuracy $Acc_{\text{PG}}$. The results through fully supervised manners are shown in the upper part, and the lower is under weak supervision. PSP denotes phrase-specific pre-training.}
		\resizebox{\linewidth}{!}{
			
			\begin{tabular}{l|cc|cc|cc}
				\hline
				& \multicolumn{2}{c|}{Nr3D++} & \multicolumn{2}{c|}{Sr3D++} & \multicolumn{2}{c}{ScanRefer++} \\
				~Method & PAG & PG  & PAG  & PG & PAG  & PG \\
				\hline
				~ScanRefer~\cite{chen2020scanrefer} & 23.8 & 30.3 & 31.4 & 36.8 &    24.4 & 33.1 \\
				~ReferIt3D~\cite{achlioptas2020referit3d} & 25.6 & 33.7 & 35.5      &  43.2     & 25.8 & 34.2  \\
				~{InstanceRefer}~\cite{yuan2021instancerefer} & 28.7 & 40.2    & 37.4   & 51.1  & 27.8  & 35.4  \\
				~SAT~\cite{yang2021sat}    & 33.2 & 48.4 & 48.2 & 62.8 & 32.3 & 40.3 \\
				~MVT~\cite{huang2022multi}    & 36.3 & 52.6 & 53.1 & 66.2 & 36.2 & 43.1 \\
				~\textbf{SAT + PSP}  &   {37.9} & {52.8}  &  {55.5}  &   {67.4} &   {40.9}   & {46.5}   \\
				~\textbf{MVT + PSP}  &   \textbf{42.1} & \textbf{56.9}  &  \textbf{61.0}  &   \textbf{72.6} &   \textbf{43.6}   & \textbf{49.1} \\\hline
				~{RandSelect}  & 8.6 & 12.8 & 13.9 & 21.7 & 10.5 & 15.0    \\
				~\textbf{SAT + PSP}  &   {13.1}   &    {20.4}   &   {19.6} &   {28.6}     & {13.3}&  {20.1}   \\
				~\textbf{MVT + PSP}  &   \textbf{17.1} & \textbf{24.7}  &  \textbf{22.5}  &   \textbf{32.2} &   \textbf{21.2}   & \textbf{24.2} \\\hline
			\end{tabular}%
			\label{tab:3dpag}%
		}
	\end{table}%

	\begin{figure}[t]
		\begin{centering}
			\includegraphics[width=1.0\linewidth]{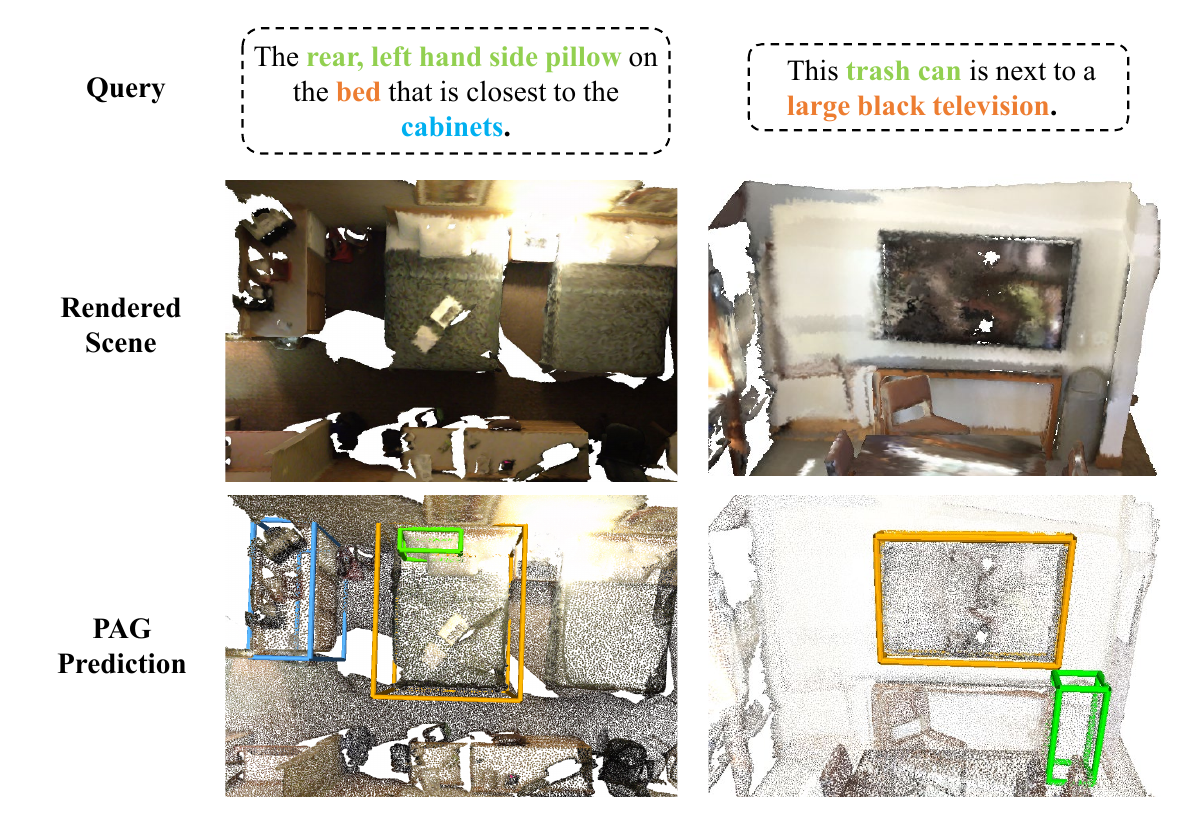}
			\caption{\textbf{Visualization Results of 3DPAG.} We show examples of 3DPAG prediction of our method. The corresponding phrase and bounding box are drawn in the same color.}
			\label{fig:fig6}
		\end{centering}	
	\end{figure}
	
	\noindent\textbf{Weakly Supervised 3DPAG-GT.}
	Since labeling phrase-level annotation is time-consuming as well, in this paper, we also explore whether we can conduct 3DPAG predictions using only 3DVG data.
	By exploiting our phrase-specific pre-training technique, our method can also conduct weakly supervised 3DPAG.
	Specifically, we first identify all object-related phrases through rule-based language parser~\cite{feng2021free}, and then construct the phrase-specific mask only for the target object (as shown in Sec.~\ref{sec:pre-training}) during the training.
	During the inference, we manually provide phrase-specific masks for non-target objects, which do not appear in training.
	In other words, we only use a subset of our 3DPAG annotations (only target phrases) in training, which can be extracted from the existing VG dataset with high precision, and evaluate on the whole dataset (both target and non-target phrases).
	We also demonstrate the results through random selection (RandSelect), and the final results are shown in the lower part of Table~\ref{tab:3dpag}.

       \noindent\textbf{3DPAG-Det.} We also evaluate our PAG task with detected proposals. Specifically, we evaluate the previous state-of-the-art in Table~\ref{tab:scanrefer}, \ie, BUTD-DETR \cite{jain2022bottom}. Although BUTD-DETR utilizes phrase-object alignment scheme during the
training, it only aligns the target object due to the lack
of annotation. During our experiment, it achieves 22.6 $Acc_{PG}@0.5$ score on ScanRefer++ dataset, and gain a boost to \textbf{32.3} with our component. It indicates that explicit supervision of phrase-object alignment is necessary for an accurate localization.
       
	\noindent\textbf{Fine-grained Grounding.}
	Figure~\ref{fig:fig6} shows the 3DPAG prediction on Nr3D++ dataset, which predicts the corresponding object of each phrase (each object is shown with the same color of the phrase). 
	Compared with traditional 3DVG, 3DPAG provides fine-grained results which would give inherent cues for explainable VG in the future.
	
	\begin{table}[t]
		\centering
		\caption{\textbf{Ablation studies on Nr3D.} The performance of using different components proposed in our paper. `PSP' and `POA Map' mean phrase-specific pre-training and POA optimization, respectively. We measure all models on Nr3D dataset.}
		\resizebox{\linewidth}{!}{
			\begin{tabular}{l|ccc|cc}
				\hline
				Method~~~ & Trans. layer & PSP & POA & SAT & MVT \\
				\hline
				Baseline    &          &              &      & 49.2  &55.1   \\
				model A   & \checkmark        &              &      & 48.3   & 54.8   \\
				model B   & \checkmark &  & \checkmark & 50.4 & 56.0 \\
				model C   & \checkmark             &     
				\checkmark        &  & 53.0    & 56.4 \\
				full model~~    &     \checkmark       &       \checkmark       &  \checkmark   & \textbf{54.4} & \textbf{59.0}    \\
				\hline
			\end{tabular}
		}
		\label{tab:abl}%
	\end{table}

	\subsection{Ablation studies}
	In this section, we investigate the effect of different components proposed in our paper, as shown in Table~\ref{tab:abl}. 
	The baseline indicates the origin SAT and MVT models, which achieve 49.2\% and 55.1\% $Acc_{\text{VG}}$ on Nr3D, respectively. 
	Purely exploiting an additional transformer layer leads to a slight performance decrease in both baselines (model A). 
	With POA optimization, model B increases the accuracy to 50.4\% and 56.0\%, which outperforms model A by 2.1\% and 1.2\%, respectively.
	Exploiting phrase-specific pre-training (model C) also improves $Acc_{\text{VG}}$ from 48.3/54.8\% to 53.0/56.4\%. 
	Both model B and model C demonstrate the effectiveness of phrase-level annotations. 
	Finally, by merging all proposed components, the full architecture obtains the highest score, \ie, 54.4\% and 59.0\%, beyond our baseline by 5.2\% and 3.9\%, respectively.

	\section{Conclusion}
	\label{sec:conclusion}
	In this paper, we introduce the task of 3D phrase aware grounding (3DPAG), which aims to localize the target object in the 3D scenes while identifying non-target ones mentioned in the sentence.
	Furthermore, we label about 227K phrase-level annotations from 88K sentences in existing 3DVG datasets.
	We exploit newly proposed datasets and introduce phrase-object alignment optimization and phrase-specific pre-training for learning object-level, phrase-aware representations.
	In the experiment, we extend previous 3DVG methods to the phrase-aware scenario and prove that 3DPAG can effectively boost the 3DVG results.
	Moreover, our enhanced models achieve state-of-the-art results on three datasets for both 3DVG and 3DPAG tasks.
	%
	%

{\small
\bibliographystyle{ieee_fullname}
\bibliography{egpaper_for_review.bbl}
}

\end{document}